\documentclass[10pt, a4paper]{article}
\pdfoutput=1
\usepackage{lrec}
\usepackage{graphicx}
\usepackage{tabularx}
\usepackage{soul}

\usepackage{amsmath}
\usepackage{amssymb}

\usepackage{booktabs}
\usepackage{float}
\usepackage{epstopdf}
\usepackage[utf8]{inputenc}

\usepackage{hyperref}
\usepackage{xstring}

\title{Counting Protests in News Articles: \\ A Dataset and Semi-Automated Data Collection Pipeline}

\name{Tommy Leung\textsuperscript{*}, L. Nathan Perkins\textsuperscript{*}}

\address{*authors contributed equally \\
         Independent Scholars \\
         Cambridge, MA, USA \\
         tommy@tommyleug.com, nathan@nathanntg.com}

\abstract{Between January 2017 and January 2021, thousands of local news sources in the United States reported on over 42,000 protests about topics such as civil rights, immigration, guns, and the environment. Given the vast number of local journalists that report on protests daily, extracting these events as structured data to understand temporal and geographic trends can empower civic decision-making. However, the task of extracting events from news articles presents well known challenges to the NLP community in the fields of domain detection, slot filling, and coreference resolution. \\
\\
To help improve the resources available for extracting structured data from news stories, our contribution is three-fold. We 1) release a manually labeled dataset of news article URLs, dates, locations, crowd size estimates, and 494 discrete descriptive tags corresponding to 42,347 reported protest events in the United States between January 2017 and January 2021; 2) describe the semi-automated data collection pipeline used to discover, sort, and review the 138,826 English articles that comprise the dataset; and 3) benchmark a long-short term memory (LSTM) low dimensional classifier that demonstrates the utility of processing news articles based on syntactic structures, such as paragraphs and sentences, to count the number of reported protest events.}

\begin{document}

\maketitleabstract

\section{Introduction}
Protests have played a notable role in social movements and political outcomes in the United States, including the Civil Rights and Anti-War movements that started in the 1960s \cite{McAdam2002,Soule2005,Andrews2006} and the rise of the Tea Party in American politics starting in 2009 \cite{Madestam2013}. Researchers that study the relationships between protests, political outcomes, and social movements often rely on time- and labor-intensive, manually coded datasets based on local newspaper reports.

Since January 2017, millions of Americans have protested about topics such as civil rights, immigration, guns, healthcare, collective bargaining, and the environment; and journalists have recorded much of this protest activity in news articles in local papers. However, for these event data to be contemporaneously useful for deeper analysis, citizens, policymakers, journalists, activists, and researchers must be able to ask structured questions of the news articles in aggregate, such as, ``How often did people protest against police brutality?''

The task of associating details with events in natural language processing is often referred to as coreference resolution or multi-slot filling and constitutes an active area of research \cite{Soderland1999,Hendrickx2009,Mesnil2015,Hakkani-Tur2016}. Multi-slot filling and coreference resolution in the context of open-ended news articles are sufficiently complex problems that they have recently inspired the definition of their own natural language processing (NLP) research task \cite{Postma2018}. To illustrate some of the specific NLP challenges related to extracting protest event details from news articles, here are example excerpts from candidate articles that contain variations of the keywords ``protest,'' ``rally,'' ``demonstration,'' or ``march'': 
\begin{enumerate}
    \item ``Protestors marched for greater gun control in Torrance.'' (one protest for gun control)
    \item ``A few dozen white nationalists held a rally; they were outnumbered by hundreds of counterprotestors.'' (two protests, one for and one against white supremacy)
    \item ``Residents in Corrales held a local demonstration to mirror the national rally in Washington, DC; rallies also occurred in New York, Boston, Seattle, Chicago, and Albuquerque.'' (seven protests, protest objective not described in this sentence)
    \item ``Teachers rallied for more funding last year, and they will do so again this year at the Capitol.'' (two protests a year apart, against systemic underfunding)
    \item ``Dow rallies 100 points.'' (zero protests)
\end{enumerate}

To help improve the resources available for extracting structured data from news stories, our contribution is three-fold. We 1) present a manually labeled dataset of news URLs corresponding to 42,347 reported protests in the United States between January 2017 and January 2021; 2) describe the semi-automated data collection pipeline used to compile the dataset nightly with a team of two researchers; and 3) benchmark a long-short term memory (LSTM) low dimensional classifier to count the number of protest events reported in a news article. 

In Section~\ref{sec:datasetStatistics}, we describe key statistics about the dataset. In Section~\ref{sec:datasetCollectionPipeline}, we describe our data collection pipeline and coding process, including key automation techniques for article discovery, similarity sorting, and classification. In Section~\ref{sec:datasetEvaluation}, we discuss the results of using the dataset to train and test a series of fully-supervised, LSTM-based neural networks to predict the number of protest events reported in a news article.

\section{Dataset Statistics}
\label{sec:datasetStatistics}

The dataset that we present in this paper\footnote{We do not have permission to directly distribute the corpus due to intellectual property rights regarding the use of news articles. Following \cite{Sharoff2005}, we release a list of web URLs (\url{https://github.com/count-love/protest-data}) and example code (\url{https://github.com/count-love/crawler}) that can be used to recreate the corpus. Given the transient nature of internet URLs, not all URLs remain available; however, where possible, multiple sources that report about the same event have been noted in the dataset.} contains 138,826 URLs corresponding to 42,347 unique protests reported in the United States between January 2017 and January 2021. In the data, we differentiate between past references to protests that have occurred and future references to protests that are planned. In total, there are 15,756 future event references and 100,963 past event references.

We first highlight other news-based civil unrest/violence datasets of social and journalistic import that may also be useful as NLP language resources for event extraction from news articles:
\begin{enumerate}
\item the Dynamics of Collective Action\footnote{\url{https://web.stanford.edu/group/collectiveaction}} database, which thoroughly documents approximately 23,000 unique, historic protest events reported in the New York Times between 1960 to 1995 and includes code fields for details such as protest targets, participation by organizations, the presence and nature of any related violence, and the number of people arrested;
\item the Event Status Corpus, which contains 4,500 news articles about civil unrest in English and Spanish with explicit annotations describing the temporal status of each reported event; \cite{Huang2016}
\item and the Gun Violence Database, which collects reports of gun violence incidents in the United States by combining an automated crawling pipeline with crowd-sourced article annotation. \cite{Pavlick2016}
\end{enumerate}
These datasets differ from the protest events dataset presented in this paper by the types of civil unrest events documented, breadth of included news sources, contemporaneity of data collection, time period covered, volume of articles, and annotation taxonomy.

\subsection{News source coverage}

\begin{figure}[h]
    \centering
    \includegraphics[width=0.48\textwidth]{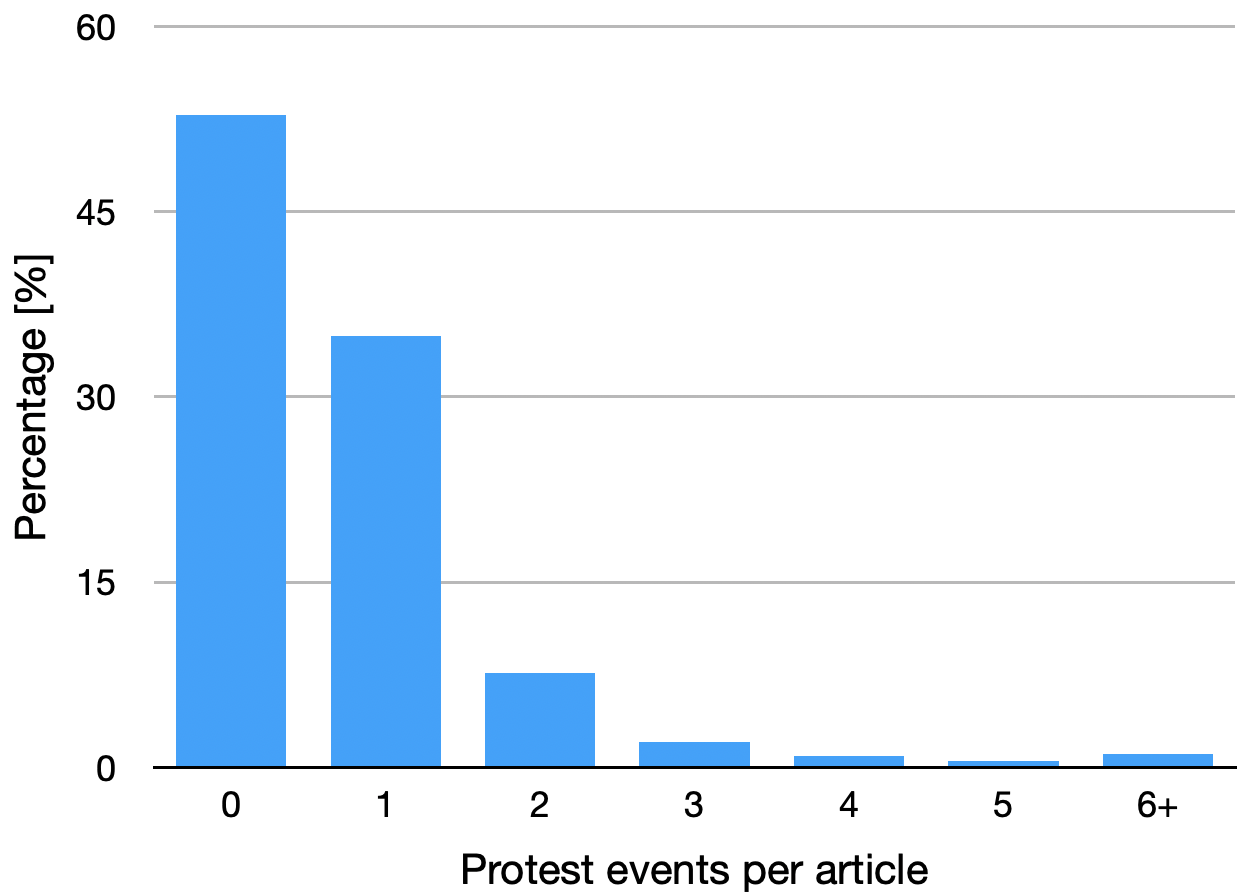}
    \caption{Distribution of the number of protest events reported per article in the dataset between January 2017 and January 2021.}
    \label{fig:distributionProtestsPerArticle}
\end{figure}

To find candidate articles for the dataset, we automatically crawled 3,410 news sources on a nightly basis. Of these news sources, 2,683 sources reported on at least one protest in the dataset; 2,363 sources reported on at least two protests; and 2,167 sources reported on at least three protests. Because the crawler automatically discovers candidate news articles using keywords, only 72,083 of the 138,826 URLs in the dataset actually describe protest events. The other 66,743 URLs are negative examples with articles about cancelled protests, stock markets, sports games, and other topics. Figure~\ref{fig:distributionProtestsPerArticle} shows a histogram of the event count distribution for all crawled articles: 95.5\% of the articles in the dataset have either zero, one, or two protest events.

In addition to enumerating all protests reported in a single news article, the dataset also enumerates all articles that report about the same protest event---this happens frequently for prominent protests, as well as when local newspapers syndicate stories from wire services such as the Associated Press or Reuters.

\subsection{Protest event tags}

\begin{table}[h]
\begin{center}
\begin{tabular}{|l|r|r|}
\hline
Category Tags & Events & Articles \\
\hline
Civil Rights &  16,130 &  31,230 \\
Collective Bargaining &  1,932 &  2,821 \\
Education &  2,305 &  4,110 \\
Environment &  1,940 &  3,209 \\
Executive &  3,750 &  7,880 \\
Guns &  3,917 &  5,262 \\
Healthcare &  2,106 &  4,290 \\
Immigration &  3,937 &  6,163 \\
International &  747 &  908 \\
Judicial &  426 &  684 \\
Legislative &  531 &  733 \\
Other &  5,156 &  7,589 \\
\hline
\end{tabular}
\caption{Protest categories as of January 31, 2021}
\label{tbl:categoryTags}
\end{center}
\end{table}

Each protest event in the dataset contains details about that protest's (1) date, (2) location, (3) number of attendees (if reported\footnote{For event attendee counts, we report the most specific, conservative estimate and choose a single source as ground truth if multiple articles provide different estimates. We interpret ``a dozen'' as 10, ``dozens'' as 20, ``hundreds'' as 100, ``a couple hundred'' as 200, etc.}), (4) article references in the dataset, and (5) tags corresponding to the category and reasons for protest. For protest tags, as of January 31, 2021, the dataset contains 494 unique tags covering the 12 general categories shown in Table~\ref{tbl:categoryTags}; 359 positions for or against a topic (e.g., ``For greater gun control'' or ``Against white supremacy''); and 123 details, including the names of national demonstrations (e.g., ``Women's March'') and common themes (e.g., ``Police'').

Every event is labeled with at least one category and one position tag. Table~\ref{tbl:categoryTags} shows the representation of category tags in the dataset as of January 31, 2021. Tags are not mutually exclusive, and an event can have multiple category, position, and detail tags.

A unique set of tags denotes a specific, semantically unique protest or protest objective---for example, ``Civil Rights; For racial justice; For greater accountability; Police'' denotes protests against racially-motivated police brutality. As of January 31, 2021, the dataset had 2,395 unique combinations of tags. Table~\ref{tbl:topTagSets} shows the top five tag sets by event count. We developed the protest taxonomy initially by manually coding events with open-ended fields. Six months after, we reviewed and binned the open-ended entries into recurring category, position, and detail tags.

\begin{table}[h]
\begin{center}
\begin{tabular}{|p{4.75cm}|r|r|}
\hline
Tag Sets & Events & Articles \\
\hline
Civil Rights; For racial justice; For greater accountability; Police & 8,353 & 16,864 \\
\hline
Guns; For greater gun control; National Walkout Day & 1,468 & 779 \\
\hline
Guns; For greater gun control & 1,208 & 1,622 \\
\hline
Civil Rights; For women's rights; Women's March & 1,021 & 1,502 \\
\hline
Healthcare; Coronavirus; Against pandemic intervention & 978 & 2,821 \\
\hline
\end{tabular}
\caption{Top five most frequent protest tag sets as of January 31, 2021. Each unique tag set denotes a distinct protest objective or set of protests.}
\label{tbl:topTagSets}
\end{center}
\end{table}

As one of our contributions in this paper, we release the dataset just described. The dataset documents protests in the United States reported by local news sources and will be made available in an archival GitHub repository.

As our primary contribution in this paper, we release the Count Love Protest Dataset just described. The dataset documents protests in the United States reported by local news sources and is available in an archival GitHub repository at \url{https://github.com/count-love/protest-data}.
    
\section{Dataset Collection Pipeline}
\label{sec:datasetCollectionPipeline}
When we started reviewing and labeling news reports in February 2017, we ran into several issues: 1) the candidate articles that we crawled often contained protest keywords even though the article itself did not describe protest events; 2) the crawler found redundant coverage about the same protest event; and 3) the crawler found exact copies of syndicated articles. We observed these issues on a recurring basis due to the volume of our review load: between February 2017 and January 2021, on average, we manually reviewed 99 articles each night with a peak of 1,296 articles. To make article discovery and review possible with a team of two researchers, we built a data collection pipeline that relies heavily on automation and broadly follows this sequence:
\begin{enumerate}
    \item each night, automatically crawl news sites to identify candidate articles with protest keywords;
    \item automatically deduplicate syndicated articles and group similar articles by topic;
    \item automatically score candidate articles to determine if the article describes at least one protest event (the domain detection task in NLP);
    \item automatically predict relevant protest details, such as the total number of reported protest events in an article, to use as suggestions during manual review; and
    \item manually read articles to label protest events.
\end{enumerate}
We describe key components of our data collection pipeline and review process in greater detail in the remainder of this section.

\subsection{News Sources and Crawler}
To find news articles that potentially contain reports of protest events, we first compiled a list of 3,410 URLs for local news sources in the United States. This list contains both home pages and local/metro sections. We assembled this list of news sources starting with newspapers linked in Wikipedia results for the query ``[state] newspapers.'' We also periodically augmented our list of new sources by adding new sources found during validation of local protest events planned as part of national advocacy efforts, such as the March for Our Lives protests for greater gun control and the Families Belong protests for more compassionate immigration policies.

We crawl each news source on a nightly basis and follow article links that contain the stem words ``march,'' ``demonstration,'' ``rally,'' or ``protest.'' Restricting our crawl to titles that contain stem words limits the scope of articles that we can automatically discover but also makes the crawl tractable. For each downloaded article, we heuristically determine the text content by parsing the article using BeautifulSoup\footnote{\url{https://www.crummy.com/software/BeautifulSoup/}} and heuristics modeled after Readability\footnote{\url{https://github.com/mozilla/readability}}. After downloading a candidate article, the text extractor scores every text element in the document object model based on characteristics such as length and punctuation and returns the most likely article text. We published the crawler's core logic at \url{https://github.com/count-love/crawler}.

\subsection{Detecting duplicates and ordering articles}
Many local news outlets will publish syndicated articles---articles that are fully, or nearly, a duplicate of another---from national services such as Associated Press and Reuters. Additionally, on any given day, news outlets often report about the same set of national topics, even if those outlets do not use syndicated content. Grouping semantically similar articles together before manual review simplifies the task of identifying duplicate news coverage of a protest event.

After crawling each night, to detect potential duplicates and group similar articles together, we apply two locality sensitive hashing techniques: a Nilsimsa-based detector (commonly also used to detect email spam) to identify similar paragraphs \cite{Damiani2004} and a document signature approach based on comparing sets of text shingle hashes \cite{Broder1997} to calculate the Jaccard similarity coefficient between two arbitrary articles \cite{Jaccard1912}. The Jaccard similarity coefficient ranges between 0 for completely different articles to 1 for completely identical articles. For pairs of articles with high Jaccard similarity coefficients, if we have already reviewed one of the two, then we run a ``diff'' operation to highlight inserted and deleted words in the new article. If a pair of articles has both a high Jaccard similarity coefficient and a similar text ``diff,'' we automatically associate all events found in the original article with the new article.

By using the Jaccard similarity coefficients as the distance metric between all pairs of unreviewed articles, we can also generalize the problem of grouping similar articles together to the traveling salesman problem \cite{Dantzig1954}: given a list of cities and the distances between all pairs of cities, what is the shortest path that visits each city once?  Empirically in our dataset, by sorting articles based on the shortest path, news articles about the same topic (even those that are not syndicated) share enough text shingles that they often appear near one another during review.

\subsection{Domain detection and event classification}
In addition to flagging duplicates and grouping similar articles together to reduce the time required for manual review, we also train and utilize fully supervised, bidirectional long short-term memory networks (BiLSTMs) \cite{Graves2005} to help suggest decisions for the tasks of domain detection (does the article actually contain a protest event) and event classification (assigning category, protest position, and detail tags to protest events). We describe these networks in greater detail in the following section.

\subsubsection{Word embeddings for machine learning}
For our neural network inputs, we rely on GloVe for word embeddings \cite{Pennington2014}. To prepare each article's text for training and scoring, we make out-of-vocabulary substitutions to the crawler text for frequent, relevant phrases that lack corresponding GloVe embeddings. To date, the most common out-of-vocabulary words are compound words related to counterprotests, such as ``counterprotested'' or ``counterdemonstration.'' Given that the subwords in these compound words do have GloVe embeddings, we hyphenate the compound words after ``counter.'' After making these substitutions, we use the spaCy\footnote{\url{https://spacy.io/}} library to tokenize and generate embedded inputs to feed into all of our neural network models, including those for domain detection and event classification.

\subsubsection{Domain detection}
To assist with the task of domain detection---determining whether a candidate article actually describes a protest, or an out-of-domain topic such as a stock market rally---we trained a fully-supervised, single-layer, 64-unit BiLSTM with a binary cross-entropy loss function and 50\% dropout to score new articles. Once scored, we manually review the titles for the lowest scoring articles each night and decide which stories to completely skip without review. Importantly, not all articles are eligible for skipping. We minimize false positives---the error of skipping articles that actually contain at least one protest event---by setting the threshold score for articles eligible for title-only review such that the false positive rate is less than 1.7\%, based on the BiLSTM classifier's receiver operating characteristic curve.


\subsubsection{Protest event classification}
Similarly, to assist with event classification---determining the category, protest position, and detail tags to assign to a particular protest event---we trained a fully supervised, single-layer, 256-unit BiLSTM with a binary cross-entropy loss function and 50\% dropout to suggest potential event tags for every candidate article. However, given that the tags that describe a protest event are generally not mutually exclusive (with the exception of opposite positional tags, such as ``For greater gun control'' and ``Against greater gun control''), the tag network has a 494 dimensional output---one dimension for each tag. This event classification network output suggests candidate tags for each article, encouraging tag reuse and a minimal protest taxonomy.

\subsection{Manual review and coding}
After automatic crawling, deduplication, sorting, domain detection, and classification each night, we manually read each candidate article to make final event coding decisions. We also manually deduplicate events by date, location proximity, and protest tags.

To date, we, as a team of two researchers, have coded the entire dataset since February 2017. Given the article review volume, we typically assign one researcher per article and jointly adjudicate ambiguities in real time. Importantly for data accuracy and consistency, the dataset released in this paper does not contain unreviewed classifier outputs. Although the machine learning models described in this section for domain detection and event classification guide our manual data entry, they do not yet perform with sufficient recall and precision for fully automatic data entry (and consequently we do not utilize them in this fashion).

\section{Low dimensional counting classifier benchmark}
\label{sec:datasetEvaluation}
In Sections \ref{sec:datasetStatistics} and \ref{sec:datasetCollectionPipeline}, we presented a protest dataset based on news articles and described the semi-automated discovery and review process used to build the dataset, including the use of standard LSTM classification networks to aid with initial article domain detection and protest event classification. In this section we benchmark the performance of a token-, sentence-, and paragraph-based low-dimensional classification network trained and tested with the dataset to automatically count the number of protest events in each article under fully supervised training. Counting the number of reported protests in a news article represents a useful step toward automatically extracting other relevant protest details because an event count can constrain the total number of subsequent slot filling tasks (e.g., if an article contains two protests, then we must also identify two dates, locations, and reasons for protest). Motivated by the distribution of protests per article in the dataset (see Figure \ref{fig:distributionProtestsPerArticle}), we recast the problem of counting protest events reported as a low dimensional classification problem where the final output can be one of four categories: zero, one, two, or more than two protest events. 

\subsection{Network structure}
\begin{figure}[h]
    \centering
    \includegraphics[width=0.48\textwidth]{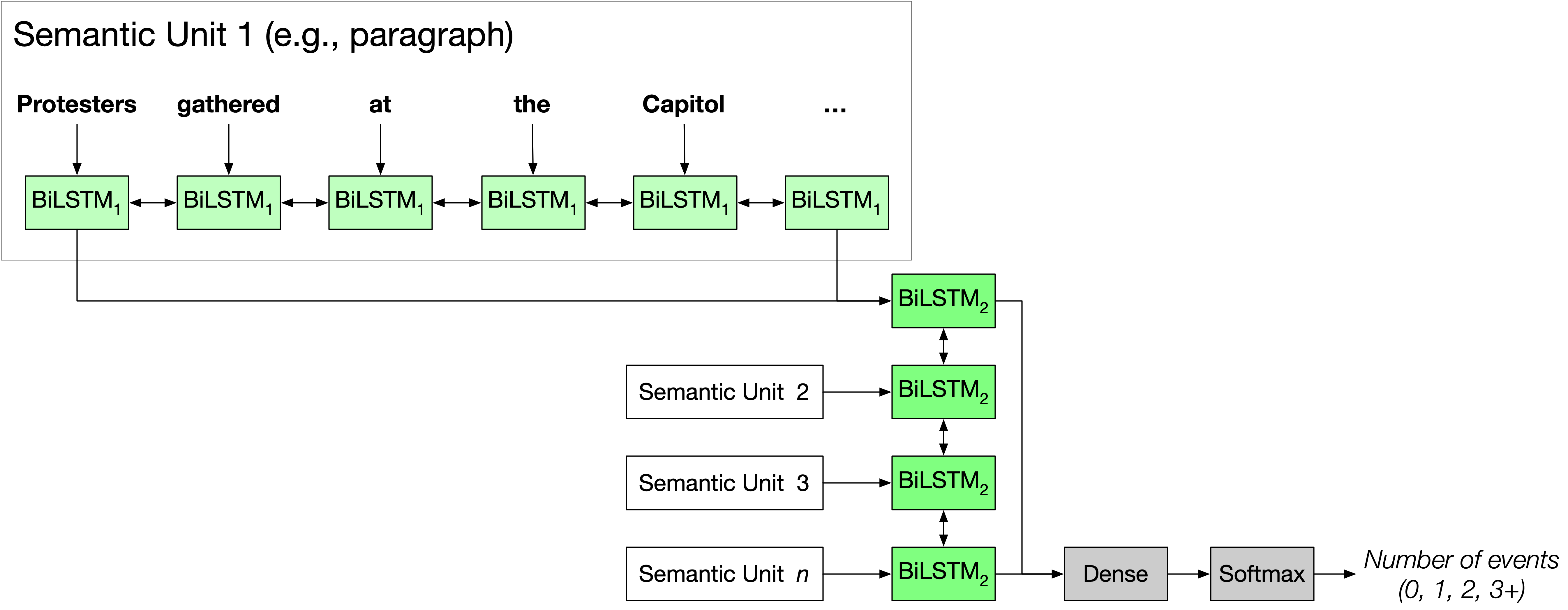}
    \caption{Given the length of most news articles, the network design for predicting the number of reported protest events first separates text into individual syntactic units, such as sentences or paragraphs, to help mitigate the long-term dependencies credit assignment problem.}
    \label{fig:model-su}
\end{figure}

To implement a low-dimensional classifier to predict the number of protest events in an article, we used a two-layer BiLSTM as shown in Figure~\ref{fig:model-su}.

Given the length of most news articles, to help mitigate the fundamental challenge of credit assignment with long-term dependencies \cite{Moniz2018}, the first layer of 128-unit BiLSTMs encodes syntactic units of text, such as sentences or paragraphs. This intermediate encoding allows the neural network to summarize sections of an article, either at the sentence or the paragraph level. For an article with $n$ syntactic units, the first layer outputs a matrix of encodings with size $n \times 256$. The second layer of 64-unit BiLSTMs operates over all $n$ syntactic units, producing an output vector of length 128. This output vector passes through a dense (fully connected) layer, followed by a final softmax layer, to predict whether the article describes 0, 1, 2, or 3+ protest events.

\begin{figure}[h]
    \centering
    \includegraphics[width=0.48\textwidth]{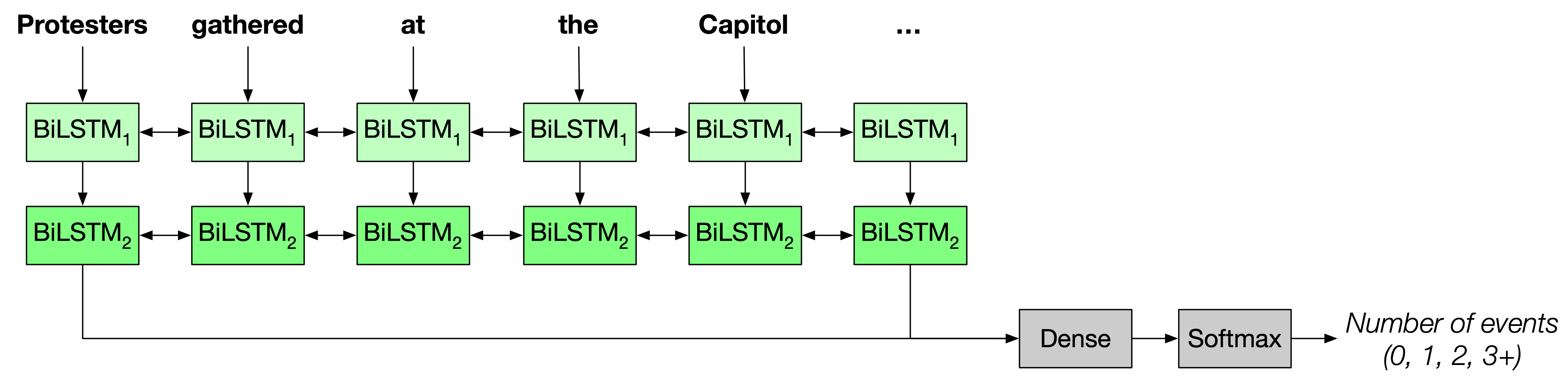}
    \caption{For benchmarking, the token-based network contains the same number of layers, units, and weights as the sentence- and paragraph-based networks, but processes articles without intermediate encoding.}
    \label{fig:model-token}
\end{figure}

To benchmark the performance gains of intermediate syntactic encoding, we also trained and tested a baseline, token-based network with no intermediate encoding. The network feeds each token through two BiLSTM layers using the same number of layers, units, and weights as the paragraph and sentence variants (see Figure~\ref{fig:model-token}).

For training and evaluation, we randomly split our corpus into three groups: 70\% for training, 15\% for validation, and 15\% for testing. We used the same document groups to test token, sentence, and paragraph syntactic units.

\subsection{Training parameters}
We trained each network using the Adam solver \cite{Kingma2015} by minimizing the categorical cross-entropy loss and passing in batches of 12 articles at a time over 10,000 training iterations. Consistent with the events per article distribution shown in Figure~\ref{fig:distributionProtestsPerArticle}, we selected articles using stratified sampling such that each batch contained 6 articles with no protest events, 4 articles with one protest event, 1 article with two protest events, and 1 article with three or more protest events. We tried training variations with larger network sizes and did not observe substantial differences in the final benchmark metrics. In addition, we measured performance on the validation dataset every 500 steps to avoid overfitting.

\subsection{Results}
After training, we evaluated the network on the withheld documents in the test set. Table~\ref{tbl:model-results} shows the results for the low-dimensional counting classification network under all three syntactic input variations: the token model with no intermediate encoding, the sentence-based model, and the paragraph-based model.

\begin{table*}[h]
    \centering
    \small
    \begin{tabular}{|l|r|r|r|r|r|r|r|r|r|}
        \toprule
        & \multicolumn{3}{c|}{Token Model} & \multicolumn{3}{c|}{Sentence Model} & \multicolumn{3}{c|}{Paragraph Model} \\
        & \multicolumn{1}{c}{P} & \multicolumn{1}{c}{R} & \multicolumn{1}{c|}{F$_1$} & \multicolumn{1}{c}{P} & \multicolumn{1}{c}{R} & \multicolumn{1}{c|}{F$_1$} & \multicolumn{1}{c}{P} & \multicolumn{1}{c}{R} & \multicolumn{1}{c|}{F$_1$} \\
        \midrule
        0 events & 93.4 & 86.1 & 89.6 & 93.6 & 88.2 & 90.8 & 95.1 & 86.1 & 90.4 \\
        1 event & 73.9 & 83.3 & 78.3 & 76.1 & 85.5 & 80.5 & 75.1 & 86.4 & 80.3 \\
        2 events & 46.7 & 27.5 & 34.6 & 54.0 & 35.0 & 42.5 & 48.6 & 38.8 & 43.2 \\
        3+ events & 38.4 & 64.6 & 48.2 & 53.7 & 70.6 & 61.0 & 53.2 & 71.6 & 61.0 \\
        \midrule
        Weighted Avg. & 81.4 & 80.4 & 80.4 & 83.3 & 83.0 & 82.9 & 83.4 & 82.5 & 82.6 \\
        \bottomrule
    \end{tabular}
    \caption{Results for the counting as a classification network under three different syntactic units of analysis: tokens, sentences, and paragraphs. Results including precision (P), recall (R), and F$_1$ score. The sentence model performed best, closely followed by the paragraph model.}
    \label{tbl:model-results}
\end{table*}

All three model variants performed best at identifying articles with zero protest events---that is, identifying articles that are outside the domain of protests. All three variants performed worst at identifying articles with two protest events, most commonly classifying these articles as containing only a single event. These errors likely reflect the similarities in representation between news articles that contain only one or two protest events, as well as the diversity of representations for reporting about two protest events (examples of which include reporting about a protest and counter-protest; a prior and current protest; or a current and future protest). 

Compared to each other, the token model slightly outperforms the sentence and paragraph models in identifying articles with no protest events. However, both the sentence and the paragraph models performed better at distinguishing between articles with one or more events, suggesting that the intermediate encoding of syntactic units can be helpful in mitigating long-term dependencies in the credit assignment problem for event counting.

In summary, the F$_1$ scores show that regardless of the syntactic unit used, recasting the counting problem to a categorization problem can be a useful simplification for counting protest events in news articles. The performance of the network with sentence- and paragraph-based intermediate encoding also suggests that larger syntactic text structures are meaningful units of analysis and can incrementally improve the performance of existing LSTM networks. 

\section{Conclusion}
Local news journalists contribute to the body of civic data by documenting community-specific events. Given the volume and frequency of their reporting work, extracting structured data about these events and enabling citizens, policymakers, journalists, activists, and researchers to query these data in aggregate can empower civic decision-making with fact-based, longitudinal analyses and narratives.

To help improve the resources available for extracting structured data from news stories, in this paper we introduced a manually labeled dataset of 42,347 protest events reported in local news sources in the United States between January 2017 and January 2021. The dataset contains metainformation about each protest, such as when, where, and why the protest occurred; a crowd size estimate; and a list of URLs to news articles that describe the protest. The dataset also contains negative examples of URLs to news articles whose titles contain protest stem words, but do not actually describe a protest event. We described the semi-automated data collection pipeline that we built to find, sort, and review news articles. Lastly, we benchmarked the performance of a low-dimensional counting classification network, trained and tested using the dataset, to count the number of protest events reported in a news article.

\section{Acknowledgements}
We thank Marek Šuppa, William Li, and Emily Prud'hommeaux for their encouragement and advice!









\section{Bibliographical References}
\label{main:ref}

\bibliographystyle{lrec}
\bibliography{arxiv}


\end{document}